\documentclass[lettersize,journal]{IEEEtran}
\usepackage{amsmath,amsfonts}
\usepackage{algorithmic}
\usepackage{array}
\usepackage[caption=false,font=normalsize,labelfont=sf,textfont=sf]{subfig}
\usepackage{textcomp}
\usepackage{stfloats}
\usepackage{url}
\usepackage{verbatim}
\usepackage{graphicx}

\usepackage{multirow}
\usepackage{booktabs}
\usepackage{xcolor}
\usepackage{float}

\def\BibTeX{{\rm B\kern-.05em{\sc i\kern-.025em b}\kern-.08em
    T\kern-.1667em\lower.7ex\hbox{E}\kern-.125emX}}
\usepackage{balance}
\begin{document}
\title{OR-NeRF: Object Removing from 3D Scenes Guided by Multiview Segmentation with Neural Radiance Fields}
\author{Youtan Yin$^{*}$, Zhoujie Fu$^{*}$, Fan Yang, Guosheng Lin
\thanks{
Corresponding author: Guosheng Lin.\\ 
\indent $^{*}$ denotes equal contribution.\\
\indent Youtan Yin, Zhoujie Fu, Fan Yang, and Guosheng Lin are with the School of Computer Science and Engineering, Nanyang Technological University (NTU), Singapore 639798 (email: youtan001@e.ntu.edu.sg, zhoujie001@e.ntu.edu.sg, fan007@e.ntu.edu.sg, gslin@ntu.edu.sg)
}}


\maketitle

\begin{abstract}
The emergence of Neural Radiance Fields (NeRF) for novel view synthesis has increased interest in 3D scene editing. An essential task in editing is removing objects from a scene while ensuring visual reasonability and multiview consistency. However, current methods face challenges such as time-consuming object labeling, limited capability to remove specific targets, and compromised rendering quality after removal. This paper proposes a novel object-removing pipeline, named OR-NeRF, that can remove objects from 3D scenes with user-given points or text prompts on a single view, achieving better performance in less time than previous works. Our method spreads user annotations to all views through 3D geometry and sparse correspondence, ensuring 3D consistency with less processing burden. Then recent 2D segmentation model Segment-Anything (SAM) is applied to predict masks, and a 2D inpainting model is used to generate color supervision. Finally, our algorithm applies depth supervision and perceptual loss to maintain consistency in geometry and appearance after object removal. Experimental results demonstrate that our method achieves better editing quality with less time than previous works, considering both quality and quantity.
\end{abstract}


\begin{figure*}[htbp]
    \centering
    \includegraphics[width=\textwidth]{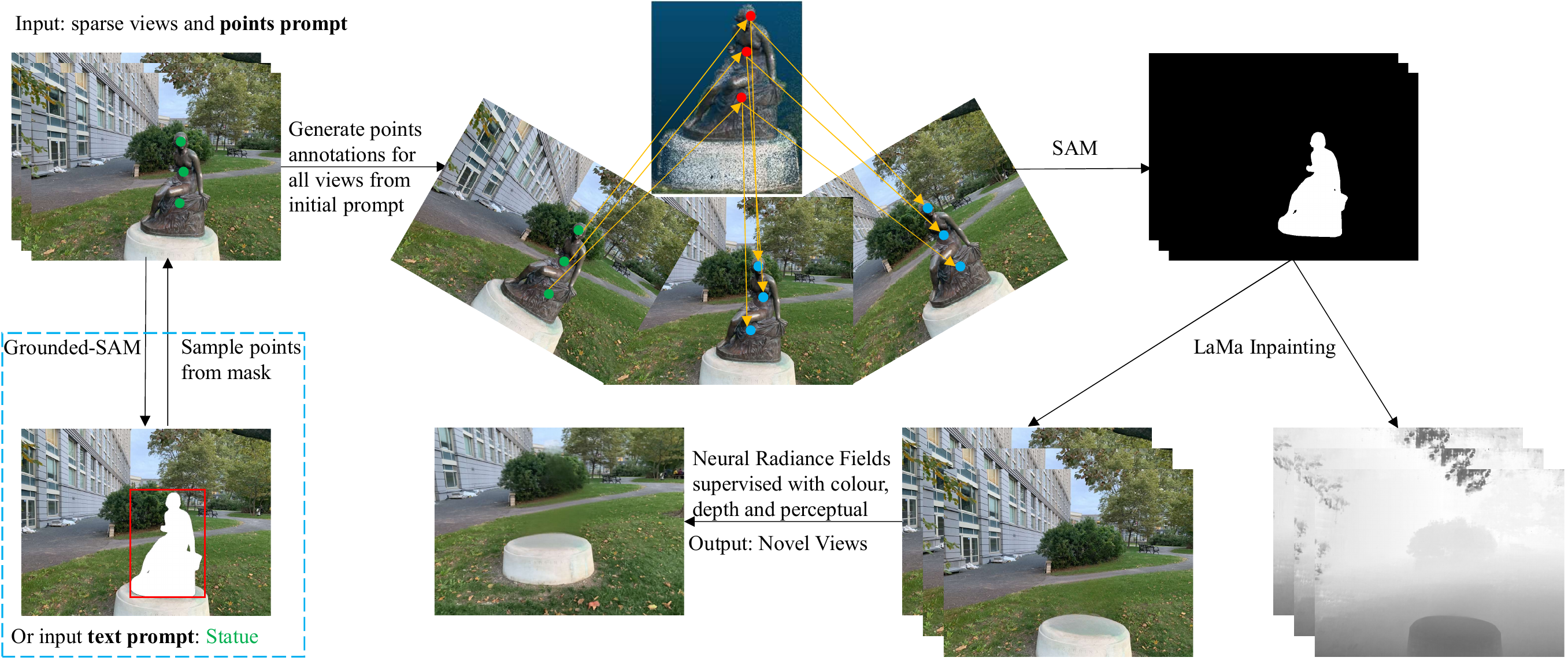}
    \caption{An overview of our OR-NeRF's framework. We start with sparse images and either points or text prompts. If a text prompt is used, we convert it into a points prompt by sampling points from the initial mask estimated using Grounded-SAM (Sec \ref{sec: Text Prompt}). Next, we propagate the points annotations to all views by projecting them from 2D to 3D point cloud and back to 2D (Sec \ref{sec: Points Prompt}). We utilize SAM to predict masks using these point annotations. LaMa is used to obtain color and depth priors. Finally, the scene after removal is reconstructed using Neural Radiance Fields supervised by color (Eq \eqref{ColourLoss}), depth (Eq \eqref{DepthLoss}), and perceptual (Eq \eqref{PerceptualLoss}) cues simultaneously (Sec \ref{sec: Scene Object Removal}).}
    \label{fig: overview}
\end{figure*}

\section{Introduction}
\label{sec: Introduction}

\IEEEPARstart{N}{eural} Radiance Fields (NeRF) \cite{mildenhallNeRFRepresentingScenes2020} has demonstrated excellent results in reconstructing 3D scenes, and recent works \cite{pengCageNeRFCagebasedNeural2022,xuDeformingRadianceFields2022,yuanNeRFEditingGeometryEditing2022,xiangNeuTexNeuralTexture2021,yangNeuMeshLearningDisentangled2022} have aimed to extend its capabilities to editing 3D scenes. One of the essential editing operations is removing objects from a 3D scene, which has garnered significant interest from the research community \cite{yangLearningObjectCompositionalNeural2021,wuObjectCompositionalNeuralImplicit2022,wederRemovingObjectsNeural2022,mirzaeiSPInNeRFMultiviewSegmentation2023,goelInteractiveSegmentationRadiance2023}. However, the practical application of this task faces several challenges. One of the most significant obstacles is the accurate localization of unwanted objects. Although it is natural for humans to identify unwanted objects, asking users to label every view is impractical. Additionally, ensuring multiview consistency and plausible content after deletion without any ground truth is not trivial.

Several works have tried to address the above problems but remain unsatisfactory. Object-NeRF \cite{yangLearningObjectCompositionalNeural2021} and ObjectSDF \cite{wuObjectCompositionalNeuralImplicit2022} decompose the NeRF training into background and objects branches, allowing them to render specified objects controlled by object ID. However, because of the lack of supervision for the removed part, neither of these works can guarantee a plausible completion at the removal area. NeRF-Object-Removal \cite{wederRemovingObjectsNeural2022} and SPIn-NeRF \cite{mirzaeiSPInNeRFMultiviewSegmentation2023} use the 2D inpainting method LaMa \cite{suvorovResolutionrobustLargeMask2022} to generate color and depth priors after deletion and reconstruct NeRF from these priors directly. Although editing quality has improved, NeRF-Object-Removal requires all views' masks to realize, while SPIn-NeRF uses a series of segmentation preliminaries \cite{haoEdgeFlowAchievingPractical2021,caronEmergingPropertiesSelfSupervised2021,zhouSurveyDeepLearning2023,zhiInPlaceSceneLabelling2021} which even involves network training to generate masks for each scene with intensive time. DFFs \cite{kobayashiDecomposingNeRFEditing2022} applies pre-trained language models \cite{radfordLearningTransferableVisual2021,caronEmergingPropertiesSelfSupervised2021} to enable text-prompt editing by training NeRF to align feature vectors extracted from language models, eliminating the need for masks. However, it has difficulty locating regions to remove if the pre-trained object detector does not work appropriately.

In this paper, we propose a novel pipeline called OR-NeRF that enables free object removal from 3D scenes using either points or text prompts on a single image, requiring less time for multiview segmentation and achieving better performance than previous methods. To spread the points prompt on a single view to other views, we introduce a point projection strategy that utilizes the COLMAP \cite{schoenberger2016vote} sparse reconstruction to find correspondences from 2D points to 3D sparse point cloud and further projects 3D points to all 2D images with camera parameters. This results in precise sparse points annotations for all scene views, which can be directly input to a recent 2D segmentation model Segment-Anything (SAM) \cite{kirillovSegmentAnything2023} to predict masks. Generated at approximately two frames per second on an RTX 3090 GPU, our algorithm outperforms previous works like SPIn-NeRF, requiring minutes. Following the approach of NeRF-Object-Removal and SPIn-NeRF, we use the 2D inpainting model LaMa to get color priors for the removal area. We develop our scene object removal algorithm using TensoRF \cite{chenTensoRFTensorialRadiance2022} as the backbone with depth supervision and perceptual loss. TensoRF is a SOTA model for improving rendering quality considering time and performance trade-offs. This approach enables us to reconstruct the 3D scene after object removal with superior editing quality compared to existing methods. Fig \ref{fig: overview} shows an overview of our OR-NeRF framework.

We evaluate our method on various datasets and analyze its performance in multiview segmentation and scene object removal through quality and quantity analyses. In summary, our contributions are (1) A novel pipeline for efficient object removal from 3D scenes, allowing for both points and text prompts on a single image, and (2) Experimental results demonstrate that our method achieves better editing quality and requires less time for multiview segmentation than previous methods, as evidenced by both quality and quantity analyses.

\section{Related Work}
\label{sec: Related Work}

\subsection{Multiview Segmentation}
\label{paragraph: Multiview Segmentation}

Though segmentation in 2D \cite{zhaofanlearning2018,leiexample2014} is well studied, multiview segmentation for 3D scenes \cite{linlifseg2023,abrarmultimodal2018} has received less attention despite its non-negligible importance for downstream applications like 3D editing. Several self-supervised methods \cite{fanNeRFSOSAnyViewSelfsupervised2022,liuUnsupervisedMultiViewObject2022} have been proposed, but they often produce inaccurate masks and have difficulty handling complex scenes. To mitigate these challenges, semi-supervised strategies \cite{wallingfordNeuralRadianceField2023,fuPanopticNeRF3Dto2D2022,zhiInPlaceSceneLabelling2021,mirzaeiSPInNeRFMultiviewSegmentation2023} have emerged that require partial annotations, or reasonable prompts from users. Semantic NeRF \cite{zhiInPlaceSceneLabelling2021} propagates partial labels to dense semantic segmentation by leveraging a few in-place annotations via predicting semantic labels with volume rendering. Like NeRF \cite{mildenhallNeRFRepresentingScenes2020}, SPIn-NeRF \cite{mirzaeiSPInNeRFMultiviewSegmentation2023} further constructs a thorough pipeline to generate masks for all views with points prompt on a single image. They use one-shot segmentation \cite{haoEdgeFlowAchievingPractical2021} to estimate an initial mask, followed by a video segmentation \cite{caronEmergingPropertiesSelfSupervised2021,suvorovResolutionrobustLargeMask2022} to generate masks for all views by treating the image sequence as a video. Finally, they refine the masks using Semantic NeRF. However, the above approaches require network training, which consumes considerable resources and does not guarantee an accurate mask, as errors can accumulate with complicated frameworks.

\subsection{Scene Object Removal}
\label{paragraph: Scene Object Removal}

NeRF has greatly facilitated the area of 3D scene editing and research \cite{baoSINESemanticdrivenImagebased2023,benaimVolumetricDisentanglement3D2022,mikaeiliSKEDSketchguidedTextbased2023,sellaVoxETextguidedVoxel2023} focuses on various editing types emerging in large numbers. Works exist for texture editing \cite{chenAUVNetLearningAligned2022,xiangNeuTexNeuralTexture2021}, geometry editing \cite{pengCageNeRFCagebasedNeural2022,xuDeformingRadianceFields2022,yuanNeRFEditingGeometryEditing2022}, and object-centred editing \cite{rematasSharfShapeconditionedRadiance2021,yuUnsupervisedDiscoveryObject2022,goelInteractiveSegmentationRadiance2023,wuObjectCompositionalNeuralImplicit2022,yangLearningObjectCompositionalNeural2021}, such as removal \cite{liuNeRFInFreeFormNeRF2022,wederRemovingObjectsNeural2022,mirzaeiSPInNeRFMultiviewSegmentation2023}, and even enabling multiple manipulations \cite{lazovaControlNeRFEditableFeature2022,kobayashiDecomposingNeRFEditing2022,wangDMNeRF3DScene2023,liuEditingConditionalRadiance2021,zhuSDFIntrinsicIndoor2023,yeIntrinsicNeRFLearningIntrinsic2023,mirzaeiLaTeRFLabelText2022,kuangPaletteNeRFPalettebasedAppearance2023}. Object-NeRF \cite{wuObjectCompositionalNeuralImplicit2022} and ObjSDF \cite{yangLearningObjectCompositionalNeural2021} decompose NeRF training into background and object branches, allowing for rendering specified objects controlled by assigned object IDs. However, they generate 'black holes' at the removal region as there is no supervision or priors for the deletion part during training. NeRF-In \cite{liuNeRFInFreeFormNeRF2022}, NeRF-Object-Removal \cite{wederRemovingObjectsNeural2022}, and SPIn-NeRF utilize the 2D inpainting method LaMa \cite{suvorovResolutionrobustLargeMask2022} to obtain priors for the removal part and directly reconstruct the scene after deletion from these priors. Though achieving better rendering quality, these methods demand high preconditions, such as annotating or generating masks for all views, which rely on expensive time costs and hardware resources. Additionally, \cite{kobayashiDecomposingNeRFEditing2022,mirzaeiLaTeRFLabelText2022,goelInteractiveSegmentationRadiance2023} combine pre-trained language models \cite{caronEmergingPropertiesSelfSupervised2021,radfordLearningTransferableVisual2021,liLanguagedrivenSemanticSegmentation2022,tschernezkiNeuralFeatureFusion2022} to enable text editing, thus bypassing the requirement for masks. Still, the rendering quality in the removal region is poor, as no algorithms are designed for learning pixel values after deletion.

\begin{figure*}[htbp]
    \centering
    \includegraphics[width=\textwidth]{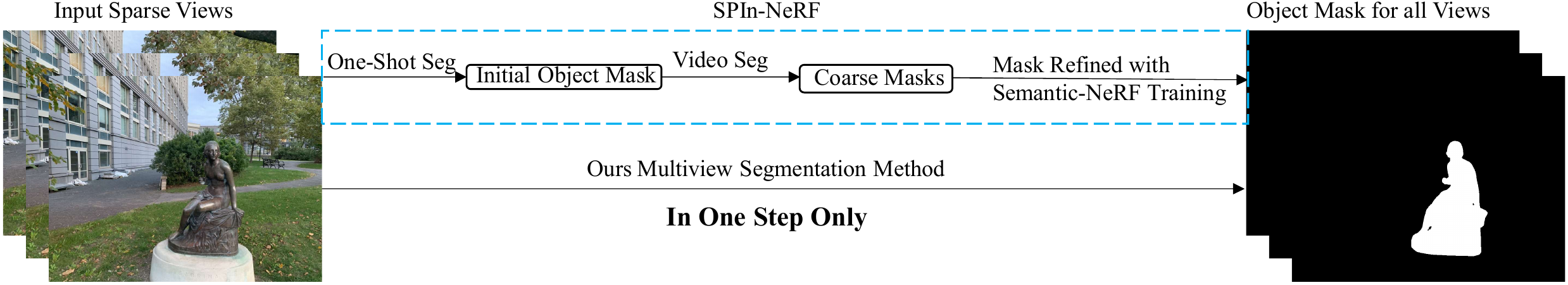}
    \caption{Comparison of mask generation between SPIn-NeRF \cite{mirzaeiSPInNeRFMultiviewSegmentation2023} (first row) and ours (second row). Our method generates masks rapidly and precisely for all views in a single step, supporting points, and text input. In contrast, SPIn-NeRF exhibits slower speed, lower accuracy, and limited support for points prompt only, requiring three steps, including network training.}
    \label{fig: mask_generation}
\end{figure*}

\section{Background}
\label{sec: Background}

\subsection{Neural Radiance Fields}
\label{sec: Neural Radiance Fields}

Given 3D location $\textbf{x}=(x,y,z)$ and 2D viewing direction $\textbf{d}=(\theta, \phi)$, NeRF models the 3D scene implicitly with an MLP network which gives a mapping function $F_{\Theta}:(\textbf{x},\textbf{d})\rightarrow(\textbf{c},\sigma)$. The output $\textbf{c}$ stands for the radiance and $\sigma$ for volume density, respectively. To optimize weights $\Theta$, the volume rendering approach is introduced as:
\begin{equation}
\label{NeRFVolumeRendering}
\begin{split}
    C(\textbf{r})&=\int_{t_{n}}^{t_{f}}T(t)\sigma(\textbf{r}(t))\textbf{c}(\textbf{r}(t), \textbf{d})dt,\\&\textrm{where}\;T(t)=\exp\left(-\int_{t_{n}}^{t}\sigma(\textbf{r}(s))ds\right)\,.
\end{split}
\end{equation}

In Eq \eqref{NeRFVolumeRendering}, $C(\textbf{r})$ represents the pixel value and is calculated by integrating the radiance value $\textbf{c}$ along the ray $\textbf{r}(t)=\textbf{o}+t\textbf{d}$ starting from the camera position $\textbf{o}$ with direction $\textbf{d}$ pointing to the pixel, within near and far bounds $t_{n}$ and $t_{f}$. The function $T(t)$ denotes the accumulated transmittance along the ray from $t_{n}$ to $t$. NeRF trains the network with the above definitions by minimizing the total squared error between rendered pixels and ground truth.

\subsection{SPIn-NeRF}
\label{sec: SPIn-NeRF}

SPIn-NeRF proposes a comprehensive pipeline for removing objects from 3D scenes. In addition to a set of sparse view images with their corresponding camera parameters, SPIn-NeRF takes a few points on one view annotated by users, indicating the unwanted objects as a prompt. With these inputs, SPIn-NeRF first combines a series of segmentation methods \cite{haoEdgeFlowAchievingPractical2021,caronEmergingPropertiesSelfSupervised2021,zhouSurveyDeepLearning2023,zhiInPlaceSceneLabelling2021} to obtain masks for all views. Then, a 2D image inpainting model LaMa \cite{suvorovResolutionrobustLargeMask2022} is used to generate color and depth priors in the mask area. The scene after deletion can be reconstructed with a modified version of vanilla NeRF from these priors directly, which adds depth supervision \cite{dengDepthsupervisedNeRFFewer2022} and perceptual loss \cite{justinPerceptulLossReal2016} to constrain the geometry and appearance consistency across different views.

In the mask generation stage, an initial mask from the single-view annotation is obtained using the one-shot segmentation \cite{haoEdgeFlowAchievingPractical2021} method. The video segmentation approach \cite{zhouSurveyDeepLearning2023,caronEmergingPropertiesSelfSupervised2021} that follows provides coarse masks for all views by wrapping images into a video sequence. Finally, the coarse masks are fine-tuned to generate proper masks for all views by fitting the Semantic NeRF \cite{zhiInPlaceSceneLabelling2021}. This procedure even requires training the Semantic NeRF from scratch to refine coarse masks obtained from \cite{zhouSurveyDeepLearning2023,caronEmergingPropertiesSelfSupervised2021}, resulting in significant costs in terms of time and hardware. Fig \ref{fig: mask_generation} shows the difference in mask generation between our pipeline and SPIn-NeRF.

\section{Method}
\label{sec: Method}

Considering a set of sparse-view images with their corresponding camera poses waiting to be edited, our method requires users to provide either points or text prompts indicating the unwanted objects for only one image. Possible prompts can be a few points marked on the object or words describing the target. To begin with, we find the masks of unwanted objects in all images. We spread the initial points prompts to all images for points input according to 3D geometry match relationships (Sec \ref{sec: Points Prompt}). While for text input, we first acquire Grounded-SAM \cite{ideaGroundedSam2023} to make an initial mask for the annotated single view followed by sampling points in this initial mask to switch text prompts to the points-prompt pattern (Sec \ref{sec: Text Prompt}).

To continue, we utilize the SAM model \cite{kirillovSegmentAnything2023} to predict masks with points prompt and use masks to guide a 2D inpainting model LaMa \cite{suvorovResolutionrobustLargeMask2022} to generate color and depth priors. Finally, we describe our object-removing strategy, guaranteeing geometry and appearance consistency across all the views (Sec \ref{sec: Scene Object Removal}). Fig \ref{fig: overview} shows an overview of our framework for removing objects from 3D scenes with points or text prompts.  

\subsection{Multiview Segmentation}
\label{sec: Multiview Segmentation}

\subsubsection{Points Prompt}
\label{sec: Points Prompt}

Suppose we have a group of $n$ images $\mathcal{I}=\{I_{i}\}_{i=1}^{n}$ and their corresponding camera parameters $\mathcal{C}=\{C_{i}\}_{i=1}^{n}$ gathered from a 3D scene. We aim to predict masks for all views $\mathcal{I}$ from only one-shot annotation. An intuitive approach to this question is to generate annotations for other images. We carefully investigate the 3D geometry matching relation in 3D scenes and find that a 2D point on a certain perspective can be spread to other views by projecting it back to 3D space and then to 2D planes under a certain camera pose. For 2D to 3D pass, we can refer to the sparse point cloud reconstructed by COLMAP \cite{schoenberger2016vote} and its projected discrete points group $\mathcal{D}=\{D_{i}\}_{i=1}^{n}$ on all 2D images. This information is represented by a certain data structure in COLMAP's sparse reconstruction as a unique one-to-one mapping, which allows us to locate points in 3D space by simply querying with 2D coordinates
. However, this introduces a new problem: finding a mapping for the user's arbitrary input is not guaranteed as the reconstruction is sparse. We can solve this question by making a query with the existing nearest points in the discrete points set $\mathcal{D}$. Finally, for the 3D to 2D reverse pass, we reproject 3D points back to the 2D plane under a certain view through its corresponding camera matrices. Now, we can spread the initial annotation provided by users to all other views safely and quickly as this algorithm utilizes 3D information, which is self-consistent and does not involve any neural network training. Only matrices computation is needed, and the algorithm can achieve a speed of about two frames per second for generating masks.

Specifically, we leverage the 3D geometry correspondence to calculate all views' annotation $\mathcal{P}_{2d}=\{P_{ij}\}_{i=1\,j=1}^{n\hphantom{=1}m}$ from the only prompt $P_{1}$ provided by users and here $P_{ij}=(x_{ij},y_{ij})$, while $m$ stands for the number of points marked in an image. With $\mathcal{P}_{2d}$, we can obtain masks $\mathcal{M}=\{M_{i}\}_{i=1}^{n}$ for all views easily from SAM model $\mathcal{F}_{S}$ by making inferences as $\mathcal{M}=\mathcal{F}_{S}(\mathcal{I},\mathcal{P}_{2d})$. To realise this, we first initialise $M_{1}$ with $\mathcal{F}_{S}(I_{1},P_{1})$. Then we acquire points $\mathcal{P}_{3d}=\{(x_{k},y_{k},z_{k})\}_{k=1}^{l}$ in 3D space by querying 2D coordinates $D_{1}^{*}=({M}_{1}\cap{D}_{1})$. Note $l$ equals the number of points in $D_{1}^{*}$ and we only refer to the points belonging to the mask ${M}_{1}$ as we need to constrain the points annotation for all views precisely match the unwanted objects. In practice, the nearest points are calculated after 3D points have been projected to 2D planes to ensure the amount and quality of prompts.

Considering the 3D to 2D situation, we begin with camera parameters. For each view $I_{i}$, the associated camera parameters $C_{i}=\{\textbf{K}_{i},\textbf{P}_{i}\}$ is composed of the intrinsics $\textbf{K}$ and extrinsics $\textbf{P}=[\textbf{R}|\textbf{t}]$. Here, the extrinsic matrix $\textbf{P}$ is represented by a $3\times3$ rotation matrix $\textbf{R}$ (camera orientation) and a $3\times1$ translation vector $\textbf{t}$ (camera position) that together transform the 3D point from the world coordinate system $P_{w}=[X_{w},Y_{w},Z_{w}]^{T}$ to the camera coordinate system $P_{c}=[X_{c},Y_{c},Z_{c}]^{T}=\textbf{R}P_{w}+\textbf{t}$. By substituting $\mathcal{P}_{3d}$ to $P_{w}$, we can switch 3D points $\mathcal{P}_{3d}$ from the world coordinate system to camera coordinate system for all views simply as $\mathcal{P}_{3d}^{*}=\{(x_{ik},y_{ik},z_{ik})\}_{i=1\,k=1}^{n\hphantom{=1}l}=\textbf{R}_{i}\mathcal{P}_{3d}+\textbf{t}_{i}$. Here $\mathcal{P}_{3d}^{*}$ denotes the camera coordinate system form. And with one little step forward:
\begin{equation}
\label{EmitDepth}
\begin{split}
\mathcal{P}_{2d}^{*}&=\{P_{ik}\}_{i=1\,k=1}^{n\hphantom{=1}l}
\\&=\{\frac{x_{ik}}{z_{ik}},\frac{y_{ik}}{z_{ik}}\}_{i=1\,k=1}^{n\hphantom{=1}l},
\\&\textrm{where}\; (x,y,z)\in\mathcal{P}_{3d}^{*}\,,
\end{split}
\end{equation}
we project 3D points $\mathcal{P}_{3d}^{*}$ back to all 2D views to get corresponding pixel coordinates $\mathcal{P}_{2d}^{*}$ in images. Now, we need to filter the number of points in each image from $k$ to $m$. To handle this issue, we spread the initial annotation $P_{1}$ to all views by performing the above 2D-3D-2D projection to $D_{1}^{*}$ similarly to get a $\mathcal{P}_{2d}^{'}=\{P_{ij}\}_{i=1\,j=1}^{n\hphantom{=1}m}$ and find the $m$ nearest points to $\mathcal{P}_{2d}^{'}$ in $\mathcal{P}_{2d}^{*}$ to construct $\mathcal{P}_{2d}$. We keep the number of points the same as the user input in each view to ensure mask quality. By far, we get all the annotations required for the prediction of SAM, and we can gain masks for all views by calling $\mathcal{M}=\mathcal{F}_{S}(\mathcal{I},\mathcal{P}_{2d})$.
    
\subsubsection{Text Prompt}
\label{sec: Text Prompt}

We leverage SAM's variety Grounded-SAM for the text prompt, which combines an object detector Grounding DINO \cite{liu2023grounding} who can handle text input. A natural way to deal with text is by asking Grounded-SAM to predict all the views' masks with the same text input. However, we observe a considerable speed drop in inference when comparing Grounding DINO to SAM. Meanwhile, Grounded-SAM can fail to handle some 'difficult' views due to Grounding DINO's limited detection ability. Therefore, we consider a two-stage strategy where we first use Grounded-SAM to obtain an initial mask for the single view and then sample points from this mask. Finally, we use the points prompt method in Sec \ref{sec: Points Prompt} to generate masks for the remaining views. This design ensures high-quality masks while minimizing computational costs.

Regarding $m$ words $\mathcal{T} = \{T_{j}\}_{j=1}^{m}$ input from the user that describe the unwanted objects. For input words sequence $\mathcal{T}$ and images $\mathcal{I}$ pairs, Grounding DINO model $\mathcal{F}_{G}$ takes the prompt $\mathcal{T}$ as labels and tries to find these labels' corresponding bounding boxes $\mathcal{B}=\{B_{ij}\}_{i=1\,j=1}^{n\hphantom{=1}m}$ in images $\mathcal{I}$ as $\mathcal{B}=\mathcal{F}_{G}(\mathcal{I}, \mathcal{T})$. As SAM is capable of two kinds of inputs, points or boxes, we can obtain the mask $M_{1}$ of unwanted objects in the user's annotated image $I_{1}$ simply by forwarding SAM with $M_{1}=\mathcal{F}_{S}(I_{1}, B_{1})$. With the one-shot mask $M_{1}$, we sample a set of $k$ points $\hat{\mathcal{P}}=\{P_{k}=(x_{k},y_{k})\}_{k=1}^{q}$ from this mask to make the problem solvable by the points prompt method (Sec \ref{sec: Points Prompt}). To implement this, we traverse the points in the mask from left to right and up to down and choose the top left, bottom right point, and center point of the mask to construct the points prompt $\hat{\mathcal{P}}$. Then, text prompt input has been converted into points prompt, and we let the algorithm used for points prompt in Sec \ref{sec: Points Prompt} generate masks for all views.

\subsection{Scene Object Removal}
\label{sec: Scene Object Removal}

Once we get object masks for all views, we can reconstruct a 3D scene without unwanted objects through Neural Radiance Fields by treating 2D inpainting priors as ground truth. Recall Sec \ref{sec: Neural Radiance Fields}, the network can be optimized by minimizing the color loss:
\begin{equation}
\label{ColourLoss}
\mathcal{L}_{c}=\Sigma_{\textbf{r}\in\mathcal{R}}||\hat{C}(\textbf{r})-C(\textbf{r})||_{2}^{2}\,,
\end{equation}
where $\mathcal{R}$ is the set of rays in each training batch, $\hat{C}(\textbf{r})$ are the ground truth and $C(\textbf{r})$ are the rendered pixels by network outputs calculated through Eq \eqref{NeRFVolumeRendering}, respectively.

However, relying solely on color loss is inadequate, as LaMa does not consider the 3D context, leading to inconsistency across different views. To address this issue, we introduce depth constraints \cite{dengDepthsupervisedNeRFFewer2022} into the training of Neural Radiance Fields. Depth values $D(\textbf{r})$ can be obtained through volume rendering easily as: 
\begin{equation}
\label{GetDepth}
\begin{split}
D(\textbf{r})&=\int_{t_{n}}^{t_{f}}T(t)\sigma(\textbf{r}(t))zdt,\ \\&\textrm{where}\;T(t)=\exp\left(-\int_{t_{n}}^{t}\sigma(\textbf{r}(s))ds\right)\,.
\end{split}
\end{equation}
where $z$ is the distance from the current 3D location to the camera position. Like RGB images, we render depth images for the original scene without deletion and use LaMa to get depth priors. Then, we add depth supervision to training as:
\begin{equation}
\label{DepthLoss}
\mathcal{L}_{d}=\Sigma_{\textbf{r}\in\mathcal{R}}||\hat{D}(\textbf{r})-D(\textbf{r})||_{2}^{2}\,,
\end{equation}
where $\hat{D}(\textbf{r})$ are the depth ground truth. We further discuss the difference between using the whole-depth image as supervision and only querying the depth in the mask area in Sec \ref{sec: ExperimentsSceneObjectRemoval}.

In addition, we recognize that depth supervision alone only enforces geometric consistency across views, while the appearance may still exhibit inconsistency. To address this, we incorporate perceptual loss \cite{justinPerceptulLossReal2016} to guide the network in learning a plausible color distribution within the masked region, matching the style of the inpainted color priors. We focus the perceptual loss specifically on the masked area. This is because color loss alone is sufficient for the non-masked area, as pixel values do not change after the deletion in this area. It is important to note that the perceptual loss is designed at the image level. In our implementation, we refer to the patch-level implementation from SPIn-NeRF, represented by the following equation:
\begin{equation}
\label{PerceptualLoss}
\begin{split}
\mathcal{L}_{p}&=\frac{1}{B}\Sigma_{\textbf{i}\in{B}}\textrm{LPIPS}(\hat{I}(\textbf{r}), I(\textbf{r}))\,,\;\\&\textrm{where}\,I(\textbf{r})=\Sigma_{\textbf{r}\in\mathcal{P}}C(\textbf{r})\,,
\end{split}
\end{equation}
and adjust the patch sampling strategy to fit a variety of data used in our Experiments (Sec \ref{sec: Datasets}). In Equation \eqref{PerceptualLoss}, we first sample a patch $\mathcal{P}$ from the mask and calculate the mean square error between the rendered pixels $I(\textbf{r})$ and the ground truth $\hat{I}(\textbf{r})$ for the pixels within the patch $\mathcal{P}$. Batch training with a size of $B$ can be employed. Finally, the training objective is to minimize the total loss $\mathcal{L}$ defined as:
\begin{equation}
\label{TotalLoss}
\mathcal{L}=a*\mathcal{L}_{c}+b*\mathcal{L}_{d}+c*\mathcal{L}_{p}\,,
\end{equation}
where $a$, $b$, and $c$ are tunable loss weights for the color, depth, and perceptual loss, respectively.

\section{Experiments}
\label{sec: Experiments}

\begin{table*}[htbp]
    \caption{\MakeLowercase{Comparison of mask generation between our method and SPIn-NeRF. The first row indicates the scene name in the SPIn-NeRF dataset, while 'points' and 'text' denote the prompts mode used, respectively.}}
    \centering
    \belowrulesep=0pt
    \aboverulesep=0pt
    \resizebox{\textwidth}{!}{
    \begin{tabular}{c|c|ccccccccc|c|cc}
    \toprule
    \multicolumn{2}{c|}{~} & 1 & 2 & 3 & 4 & 7 & 9 & 10 & 12 & trash & Mean & SPIn-NeRF \\
    \midrule
    \multirow{2}*{points} & acc$\uparrow$ & 99.80 & 99.82 & 99.73 & 99.79 & 99.81 & 99.78 & 99.87 & 99.30 & 99.51 & 99.71\textcolor{green}{$\uparrow$} & 98.91 \\ 
    ~ & IoU$\uparrow$ & 96.77 & 96.47 & 97.48 & 98.50 & 97.43 & 96.29 & 95.47 & 91.73 & 88.68 & 95.42\textcolor{green}{$\uparrow$} & 91.66 \\
    \midrule
    \multirow{2}*{text} & acc$\uparrow$ & 99.81 & 99.82 & 99.73 & 99.80 & 99.81 & 99.78 & 99.86 & 99.25 & 99.51 & 99.71\textcolor{green}{$\uparrow$} & 98.91  \\ 
    ~ & IoU$\uparrow$ & 96.81 & 96.51 & 97.47 & 98.51 & 97.43 & 96.41 & 95.41 & 91.19 & 88.64 & 95.38\textcolor{green}{$\uparrow$} & 91.66 \\
    \bottomrule
    \end{tabular}}
    \label{tab: multiview segmentation}
\end{table*}

\begin{figure*}[htbp]
    \centering
    \includegraphics[width=\textwidth]{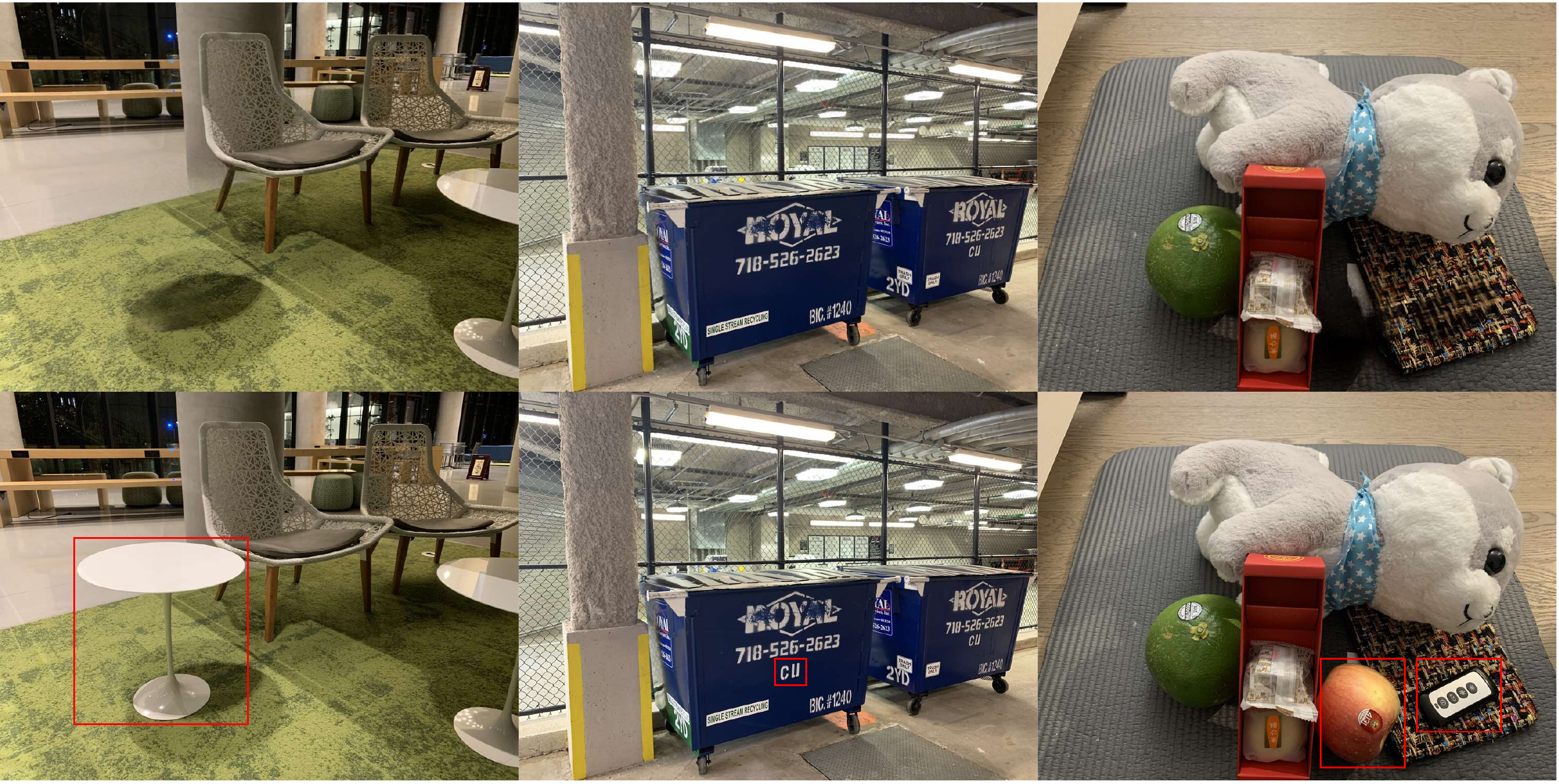}
    \caption{Editing results of our OR-NeRF method demonstrating various examples. Please zoom in to observe better.}
    \label{fig: sor_all}
\end{figure*}

\subsection{Datasets}
\label{sec: Datasets}

We select 12 scenes from various commonly used 3D reconstruction datasets, including NeRF LLFF data, IBRNet data \cite{wang2021ibrnet}, and LLFF real-world data \cite{mildenhall2019llff}. Our scene selection aims to cover a wide range of scene variations and different types of removal operations, such as slogans, providing a high degree of flexibility. Since the reconstruction datasets do not provide ground truth for evaluation, we incorporate the SPIn-NeRF dataset, which includes human-annotated object masks and scene capture after object removal. We use all 10 scenes from the SPIn-NeRF dataset to evaluate the quality of multiview segmentation. To evaluate scene object removal's performance, we select 8 scenes, excluding two duplicate scenes, to ensure a diverse layout of the objects. To conclude, we conducted experiments on 20 scenes, comprehensively evaluating our OR-NeRF pipeline.

\subsection{Metrics}
\label{sec: Metrics}

We adopt the evaluation metrics commonly used in segmentation tasks, including pixel-wise accuracy (Acc) and intersection over union (IoU), to assess the performance of our multiview segmentation algorithm. We report peak signal-to-noise ratio (PSNR), a widely used 3D reconstruction metric for the scene object removal component. Additionally, we include two metrics used by SPIn-NeRF \cite{mirzaeiSPInNeRFMultiviewSegmentation2023}: the learned perceptual image patch similarity (LPIPS) \cite{zhangUnreasonableEffectivenessDeep2018} and the Fr\'{e}chet inception distance (FID) \cite{heuselGANsTrainedTwo2017}. These metrics compare the similarity between the ground-truth data and the rendering outputs produced by our method.

\subsection{Experiments Settings}
\label{sec: Experiments Settings}

\subsubsection{Multiview Segmentation}
\label{sec: ExperimentsSettingsMS}

We conduct experiments using points and text prompts on our selected scenes and evaluate the results using metrics stated in Sec \ref{sec: Metrics}. Since the implementation details of multiview segmentation were not explicitly provided in the SPIn-NeRF paper, we directly utilize the metrics mentioned in their paper. However, it should be noted that the paper does not specify which scenes were used for calculating these metrics. Therefore, we compare the performance of SPIn-NeRF with our scene-average results. Subsequently, we utilize the masks generated from the points prompt for all subsequent experiments.

\subsubsection{Scene Object Removal}
\label{sec: ExperimentsSettingsSOR}

We conducted experiments on all 20 scenes with ours and the SPIn-NeRF methods. Both vanilla NeRF and TensoRF architectures are tested with our method's implementation. We follow the implementation of the SPIn-NeRF to reproduce their results. For NeRF and TensoRF, we train the original scenes to render depth maps instead of disparity maps used in SPIn-NeRF. This decision is made to avoid errors from dividing by zero when calculating disparities.

\subsection{Multiview Segmentation}
\label{sec: ExperimentsInteractiveMultiviewSegmentation}

\begin{figure*}[htbp]
    \centering
    \includegraphics[width=\linewidth]{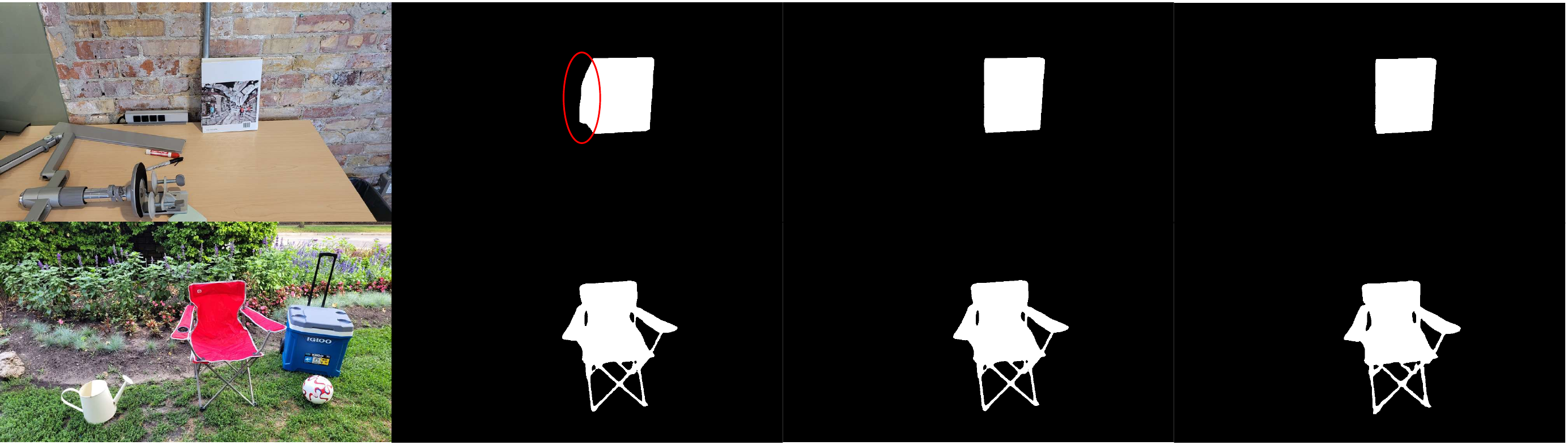}
    \caption{Mask generation results of our OR-NeRF. The figure shows the masks generated for two scenes: 'book' (up) and '12' (below) from the SPIn-NeRF dataset. From left to right are the original image, ground truth mask, masks generated with points, and text prompt.}
    \label{fig: ms}
\end{figure*}

\begin{table}[htbp]
    \caption{Comparison of time consumption between our method and SPIn-NeRF \cite{mirzaeiSPInNeRFMultiviewSegmentation2023}. Statics for SPIn-NeRF are borrowed from their paper directly.}
    \centering
    \belowrulesep=0pt
    \aboverulesep=0pt
    \begin{tabular}{c|ccc}
    \toprule
    ~ & One-shot Seg & Video Seg & 3D Fit \\
    \midrule
    SPIn-NeRF & \textless 1 sec & \textless 1 min & 3-6 min \\ 
    \midrule
    Ours & \multicolumn{3}{c}{\textless 1 min} \\
    \bottomrule
    \end{tabular}
    \label{tab: multiview segmentation time}
\end{table}

Table \ref{tab: multiview segmentation} compares mask generation between our method and SPIn-NeRF. Our approach outperforms SPIn-NeRF regarding accuracy and IoU. SPIn-NeRF's mask generation process involves a complex pipeline that introduces errors at each step and requires significant time and hardware resources. In contrast, our method leverages the simplicity of SAM and involves minimal matrix calculations. Consequently, our multiview segmentation algorithm delivers superior-quality results in less time. Table \ref{tab: multiview segmentation time} shows our estimated time for mask generation compared to SPIn-NeRF.

Note that we have excluded the 'book' scene from the average calculation. This decision was made because we have identified inaccuracies in the ground truth labels for this particular scene, as evident from Fig \ref{fig: ms}. Furthermore, as depicted in Fig \ref{fig: ms}, our segmentation results exhibit precise coverage of the target objects with intricate details, such as the crossing chair legs in the '12' scene. However, it should be noted that there is a minor flaw in the 'trash' scene where our masks fail to cover all areas of the trash cans, explaining the low metrics in Table \ref{tab: multiview segmentation}. This does not significantly affect the subsequent experiments if refined with our strategy.

\subsection{Scene Object Removal}
\label{sec: ExperimentsSceneObjectRemoval}

\begin{table*}[htbp]
    \caption{Experiment results on scene object removal. The first row indicates the method name, while abbreviations in the second row indicate loss modules. 'dir' denotes training Neural Radiance Fields with LaMa priors directly, 'dp' denotes partial depth, 'da' denotes all depth, and 'lpips' denotes the use of perceptual loss. Notably, perceptual loss is always applied with all-depth supervision enabled.}
    \centering
    \belowrulesep=0pt
    \aboverulesep=0pt
    \resizebox{\textwidth}{!}{
    \begin{tabular}{c|cccc|ccc|ccc}
    \toprule
    ~ & \multicolumn{4}{c|}{Ours-NeRF} & \multicolumn{3}{c|}{Ours-TensoRF} & \multicolumn{3}{c}{SPIn-NeRF} \\
    ~ & dir & dp & da & lpips & dir & da & lpips & dir & da & lpips \\
    \midrule
    PSNR$\uparrow$ & 14.04 & 14.04 & 14.16 & 14.16 & 13.93 & 14.04 & 14.03 & \textbf{14.85} & 14.82 & 14.83 \\
    FID$\downarrow$ & 61.11 & 65.21 & 64.71 & 58.15 & \textbf{53.28} & 64.29 & 59.74 & 70.02 & 70.07 & 67.26 \\
    LPIPS$\downarrow$ & 0.6834 & 0.6893 & 0.7022 & 0.6763 & 0.6370 & 0.6494 & \textbf{0.6273} & 0.6810 & 0.6752 & 0.6506 \\
    \bottomrule
    \end{tabular}}
    \label{tab: object removal}
\end{table*}

\begin{figure*}[htbp]
    \centering
    \includegraphics[width=\linewidth]{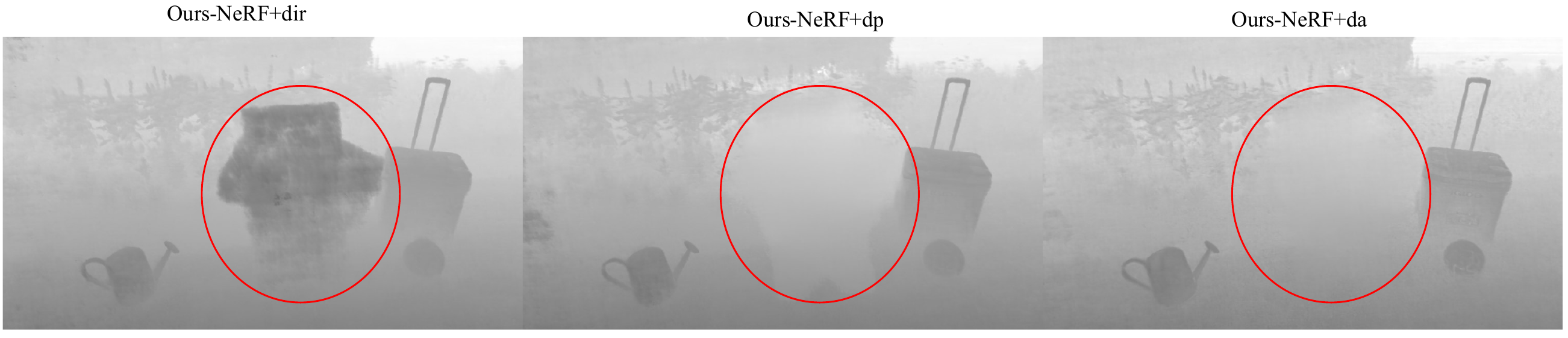}
    \caption{The effect of depth supervision. We can see from the figure that either without depth supervision (left) or training with partial depth (middle) leads to geometry inconsistency. While supervised by all-depth images (right) convergent to a consistent result.}
    \label{fig: sor_single_depth}
\end{figure*}

\begin{figure}[htbp]
    \centering
    \includegraphics[width=\linewidth]{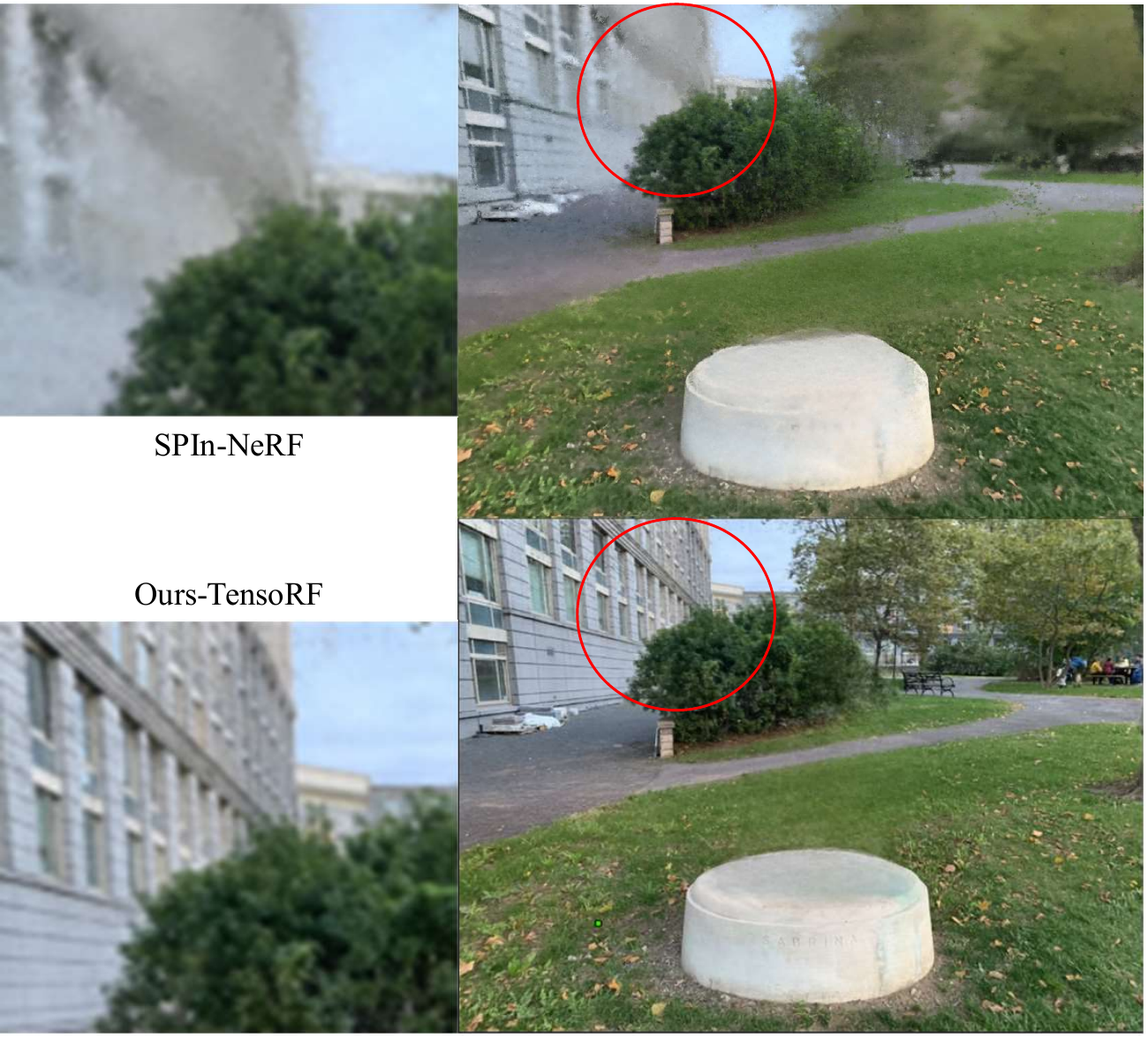}
    \caption{Comparison of overall rendering quality between SPIn-NeRF (top) and Ours-TensoRF (bottom). We can see blurry from SPIn-NeRF compared with our clear details.}
    \label{fig: sor_single_overall}
\end{figure}


\begin{figure}[htbp]
    \centering
    \includegraphics[width=\linewidth]{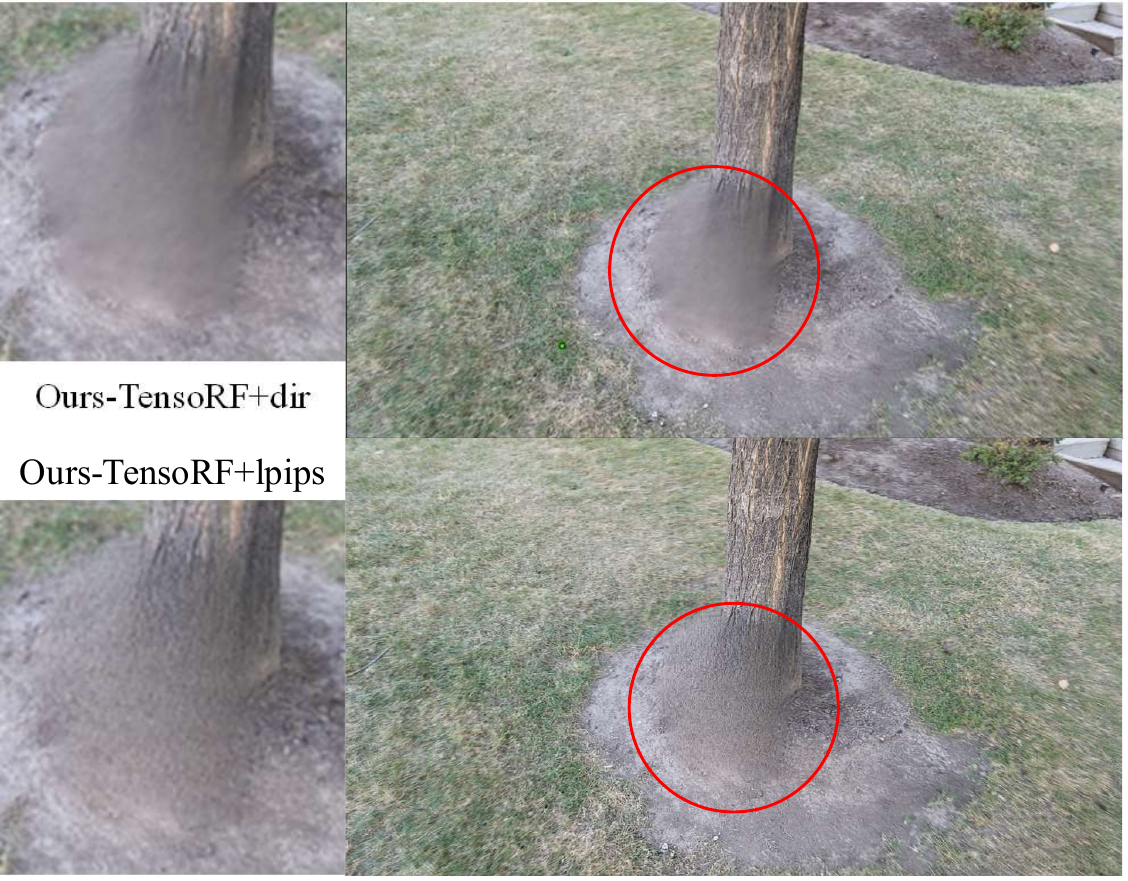}
    \caption{The effect of perceptual loss. The top is Ours-TensoRF trained directly, and the bottom is Ours-TensoRF with perceptual loss. We can see from the figure that this loss has some influence but is still unsatisfactory.}
    \label{fig: sor_single_lpips}
\end{figure}

\subsubsection{Quantity}
\label{sec: ExperimentsSORQuantity}

Table \ref{tab: object removal} presents our results for scene object removal. Regarding overall rendering quality, Ours-NeRF exhibits a superior FID compared to SPIn-NeRF but performs worse regarding PSNR and LPIPS. On the other hand, Ours-TensoRF outperforms SPIn-NeRF in terms of FID and LPIPS scores but has a weakness in PSNR. Analyzing the impact of the loss models, it appears that the additional components for training Neural Radiance Fields do not have a significantly positive effect. Ours-NeRF and ours-TensoRF exhibit a similar pattern where depth supervision and perceptual loss increase PSNR but show no positive influence on FID and LPIPS.

Interestingly, SPIn-NeRF behaves somewhat differently: removing perceptual loss and depth supervision from the SPIn-NeRF pipeline results in a subtle increase in PSNR compared to the original version. However, the FID and LPIPS scores demonstrate that the add-ons improve SPIn-NeRF performance. While the results involve complex numbers, we adopt Ours-TensoRF with perceptual loss as it performs best overall. Although Table \ref{tab: object removal} does not provide strong evidence for the efficacy of depth supervision and perceptual loss, we will discuss the real significance of these add-ons in the following section.


\subsubsection{Quality}
\label{sec: ExperimentsSORQuality}

We first compare the three methods' overall rendering quality in this part. Ours-NeRF and Ours-TensoRF produce clear outputs, while SPIn-NeRF suffers from blurry due to the noisy disparity maps, which provide inaccurate geometry supervision. This can be observed by Fig \ref{fig: sor_single_overall}.

Next, we discuss the impact of depth supervision. Although widely used in training, there is a lack of exploration of the difference between using the entire depth image as supervision and only applying depth loss in the masked area. Fig \ref{fig: sor_single_depth} indicates that full-depth supervision is necessary and irreplaceable, as both partial depth and direct training settings in all three architectures produce inconsistent depth results, resulting in different extents of restoring removed objects. However, it is worth noting that the depth loss does not show a visible difference in the rendered views, which aligns with the metrics presented in Table \ref{tab: object removal}.

Moving on to the perceptual loss aspect, we conclude from Fig \ref{fig: sor_single_lpips} that this loss has a positive effect but falls short of guaranteeing a plausible completion for the masked area. This also explains the relatively ineffective metrics in Table \ref{tab: object removal}, as our results exhibit a significant gap with the ground truth. Finally, part of our editing results are displayed in Fig \ref{fig: sor_all}.

\section{Conclusions and Discussions}
\label{sec: Conclusions and Discussions}

This paper presents a novel pipeline OR-NeRF for object removal from 3D scenes, requiring only points or text prompts on a single view. We emphasize the advantages of our method in terms of rendering quality and time efficiency. Potential limitations exist due to the inpainting model's capability and more robust 2D image inpainting techniques, such as diffusion \cite{zhangAddingConditionalControl2023,haqueInstructNeRF2NeRFEditing3D2023,rajDreamBooth3DSubjectDrivenTextto3D2023,pooleDreamFusionTextto3DUsing2023,zhouSparseFusionDistillingViewconditioned2023,singerTextTo4DDynamicScene2023,shidongdifffashion2023} based methods can be applied to achieve more plausible completions after object removal. 

\bibliographystyle{IEEEtran}
\bibliography{tmm.bib}

\end{document}